\pdfoutput=1

\documentclass[11pt]{article}

\usepackage{acl}

\usepackage{times}
\usepackage{latexsym}
\usepackage{caption}
\usepackage{subcaption}

\usepackage[T1]{fontenc}
\usepackage{amsmath}
\usepackage{amssymb}
\usepackage{enumitem}
\usepackage{float}
\usepackage{graphicx}
\usepackage{booktabs}

\usepackage[utf8]{inputenc}

\usepackage{microtype}

\usepackage{inconsolata}

\usepackage{graphicx}

\title{MLSD: A Novel Few-Shot Learning Approach to Enhance Cross-Target and Cross-Domain Stance Detection} 

\author{Parush Gera \\
  University of South
Florida \\
  Tampa \\ 
  United States \\
  \texttt{parush@usf.edu} \\\And
  Tempestt Neal \\
  University of South
Florida \\
  Tampa \\
 United States \\
  \texttt{tjneal@usf.edu} \\}

\begin{document}
\maketitle
\begin{abstract}

\end{abstract}

We present the novel approach for stance detection across domains and targets, \textbf{M}etric \textbf{L}earning-Based Few-Shot Learning for Cross-Target and Cross-Domain \textbf{S}tance \textbf{D}etection (\textbf{MLSD}). MLSD utilizes metric learning with triplet loss to capture semantic similarities and differences between stance targets, enhancing domain adaptation. By constructing a discriminative embedding space, MLSD allows a cross-target or cross-domain stance detection model to acquire useful examples from new target domains. We evaluate MLSD in multiple cross-target and cross-domain scenarios across two datasets, showing statistically significant improvement in stance detection performance across six widely used stance detection models. 

\section{Introduction}

Stance detection—the task of identifying an author's stance toward a specific target—has garnered significant attention in the research community, particularly gaining recognition during Task 6 (stance detection on social media posts) of the 2016 International Workshop on Semantic Evaluation (SemEval 2016) \cite{mohammad2016semeval}. 
In stance detection, a classification model is trained to classify an author's stance toward some target, such as a person, entity, or organization (see Table \ref{tab:stance_example}). Typically, an author's stance is classified as \textit{in favor of}, \textit{against}, or \textit{neutral}/\textit{neither}, although more recent works have introduced categories such as \textit{support}, \textit{agree/disagree}, \textit{refute}, \textit{discussion}, \textit{commenting}, and \textit{unrelated} \cite{conforti2020willtheywontthey}. In the vanilla stance detection problem, the target of interest in the training data is also the target in the test samples, meaning that learning across targets or domains is not required. However, \textit{cross-target} and \textit{cross-domain} stance detection have seen significant growth \cite{augenstein2016stance, xu-etal-2018-cross, 10.1145/3331184.3331367, conforti2020willtheywontthey}, with important implications for improving model adaptability and generalization across varied contexts and domains \cite{hardalov-etal-2021-cross}.

\begin{table}[ht!]
\centering
\caption{Example of stance from a text sample in the SemEval 2016 dataset \cite{mohammad2016semeval}.}
\resizebox{\columnwidth}{!}{
\begin{tabular}{p{2cm}p{5cm}c}
\hline
\textbf{Target} & \textbf{Text} & \textbf{Stance} \\ \hline

Legalization of Abortion & Whether someone wants to have children or not should be completely up to the person carrying that pregnancy. & \texttt{FAVOR} \\ 
\hline
\end{tabular}}

\label{tab:stance_example}
\end{table}

Cross-target stance detection (CTSD) involves training a model on a specific \textit{source} target, and using that model to predict stance expressed on a different \textit{destination} target. In this case, the source and destination data are assumed to belong to the same contextual domain \cite{xu-etal-2018-cross, augenstein2016stance}. In cross-domain stance detection (CDSD), however, the source and destination targets belong to different contextual domains. For example, the SemEval 2016 dataset included targets  \textit{Donald Trump} and \textit{Hillary Clinton}, which can be considered under the same contextual domain of \textit{American Politics}; similarly, targets  \textit{Feminist Movement} and \textit{Legalization of Abortion} were categorized under the \textit{Women’s Rights} domain by existing studies \cite{xu-etal-2018-cross}. An example of CTSD would be detecting stance toward \textit{Hillary Clinton}-based samples using a model trained on \textit{Donald Trump}-based samples, while CDSD would involve predicting stance toward \textit{Hillary Clinton} using a model trained on \textit{Feminist Movement}.

Despite the growth of research in CTSD and CDSD, significant challenges remain that limit further advancements. For example,  existing datasets often contain a wide variety of targets and domains \cite{gera-neal-2022-comparative}, making it impractical to train a separate model per target. Further, as new targets and domains are introduced, dataset annotation can also be very expensive and time-consuming. 
In addition, the disparity between source and destination domains often results in poor model transferability \cite{conforti2020willtheywontthey}. Current approaches (e.g., \cite{commonsense_stance, prompt_based_stance, semi_supervised_stance, conditional_generation_stance} ) often rely on external knowledge sources, such as knowledge graphs, Wikipedia data, target explanations, or large language models, to enhance generalization across targets and domains. However, these methods face difficulty in capturing domain-specific nuances or rely on the availability of external resources. Some techniques, such as SiamNet \cite{siamnet} and the bi-conditional model \cite{augenstein2016stance} \cite{conforti2020willtheywontthey}, have also been shown to struggle with domain-specific language, contextual differences, and target-related details that are not well captured by external knowledge alone. Some models may also be prone to overfitting on the specific characteristics of the training data, as demonstrated by \cite{sultan-etal-2022-overfit}, leading to poor transferability when applied to new targets or domains. 

One potential solution is few-shot learning, where a small number of representative examples from the destination target are used to fine-tune a stance detection model pre-trained on a different source target. Instead of relying on large datasets or external knowledge, which are often costly and domain-specific, few-shot learning would instead select the most informative examples that encapsulate relevant contextual knowledge to assist in training. Few-shot learning has been used in stance detection, primarily to to reduce the reliance on large labeled datasets and enable models to perform well with limited examples \cite{10098547, 10.1145/3543507.3583250, prompt_based_stance}. However, these approaches do not seek to effectively support  generalization across domains, which we propose as a novel paradigm for improving model performance in CTSD and CDSD. 


In this paper, we introduce a new method called \textbf{MLSD} (\textbf{M}etric \textbf{L}earning-Based Few-Shot Learning for Cross-Target and Cross-Domain \textbf{S}tance \textbf{D}etection) to improve CTSD and CDSD. Our method uses \textit{metric learning} to measure the similarity between text samples from different targets, selecting a small number of representative samples from the destination data (the ``few shots''). These selected samples are then used to further train a stance classification model, which was initially trained on a different target, to accurately predict the stance in the new destination domain. We demonstrate that our approach improves performance in both CTSD and CDSD across two datasets, which include four distinct targets and two distinct domains, using six state-of-the-art stance detection models. \footnote{Our code is available at: https://github.com/parushgera/mlsd-few-shot} Our contributions include: 
\begin{enumerate}[noitemsep, topsep=0pt]

\item \textbf{Similarity-based few-shot selection strategy:} We develop a metric-based approach to select contextually relevant samples from a destination domain for few-shot learning.

\item \textbf{Few-shot learning for cross-domain adaptation:} We propose a few-shot learning framework that improves both cross-target and cross-domain stance detection performance.



\end{enumerate}

The paper is outlined as follows. Section 2 reviews the related work; Section 3 details the proposed methodology; Section 4 presents the experiments; Section 5 evaluates the results, Section 6 concludes the paper with key findings and contributions, Section 7 discusses study limitations and Section 8 discusses the Ethical Implication of the study.
\section{Related Work}

\label{sec:related_work}


\subsection{Cross-Target Stance Detection}

Several authors have investigated CTSD using a variety of approaches. For example,  \citet{augenstein2016stance} studied CTSD by developing a conditional encoder model for predicting stance on unseen targets (e.g., Donald Trump) that were not included in the training set of the SemEval 2016 dataset. \citet{xu-etal-2018-cross} explored CTSD on domain-related targets, where samples from the source target were leveraged to learn domain-specific aspects that aid stance detection on the test data. However, there approaches are limited to scenarios where the source and destination targets belong to the same contextual domain. \citet{10.1145/3331184.3331367} investigated CTSD by leveraging labeled data from a source target to adapt models for a destination target using shared latent topics as transferable knowledge. However, this approach relies heavily on identifying shared latent topics between the targets, which may not be available or relevant across all target pairs.  \citet{10.1145/3442381.3449790}  proposed a method that builds target-adaptive pragmatics dependency graphs  to capture the roles words play in expressing stance based on the target. Their approach constructed separate graphs for both target-specific and cross-target contexts, using a graph-aware model with Graphical Convolutional Network blocks to adapt the model for unseen targets. While effective, this method relies on complex graph structures and the identification of target-dependent word roles, which may not be applicable across all target types.

\subsection{Cross-Domain Stance Detection}

CDSD poses a greater challenge by requiring the model to generalize across different contextual domains. \citet{conforti2020willtheywontthey} investigated SiamNet \cite{siamnet} and BiCond \cite{augenstein2016stance}  architectures for CDSD and found poor generalization across Entertainment and Healthcare domains in the WT-WT dataset \cite{conforti2020willtheywontthey}. \citet{hardalov-etal-2021-cross} explored CDSD for unseen targets and labels, proposing a framework that combined domain adaptation techniques, such as a mixture of experts and domain-adversarial training, with label embeddings. However, the complexity of this method requires additional label information, which may not always be available.  \citet{arakelyan-etal-2023-topic} proposed Topic Efficient Stance Detection, which uses a topic-guided diversity sampling method to create a data-efficient training set and a contrastive learning objective to fine-tune the stance classifier. Their approach addresses class imbalances across topics and evaluates both in-domain  and out-of-domain performance. However, their method depends on identifying and utilizing topic information.

\subsection{Few-Shot Learning in Stance Detection}

\citet{commonsense_stance} examined a more realistic stance detection scenario in zero-shot and few-shot settings by incorporating commonsense relational knowledge to enhance generalization and reasoning abilities. However, their work is confined to standard stance detection and does not address CTSD or CDSD. \citet{prompt_based_stance} proposed a prompt-based fine-tuning approach for stance detection using pre-trained language models to address both few-shot learning and the lack of contextual information for targets. While their approach relies on prompt design and verbalizers to capture target-context relationships, our method leverages similarity-based learning to optimize few-shot selection and improve performance across diverse targets and domains without the need for hand-crafted prompts. Furthermore, using large language models for prompt-based fine-tuning can be computationally expensive, whereas our MLSD approach achieves effective generalization and sample selection without the added cost and complexity of LLMs. \citet{conditional_generation_stance} reformulated zero-shot and few-shot stance detection as a conditional generation task using denoising from partially-filled templates, incorporating target prediction and Wikipedia knowledge to enhance performance. Their method relies heavily on external knowledge sources, whereas our MLSD framework focuses on improving generalization across both targets and domains through similarity-based selection, eliminating the need for external knowledge bases. \citet{semi_supervised_stance} proposed a target-aware semi-supervised framework for few-shot stance detection, utilizing contrastive learning and consistency regularization to leverage both labeled and unlabeled data, focusing on standard stance detection. However, their approach relies on randomly selecting few-shot samples, while our MLSD method selects samples based on learned similarities and consistently outperforms random selection, particularly in cross-target and cross-domain scenarios, without the need for semi-supervised techniques. 


\section{Proposed Approach: MLSD}
\label{sec:Methodology}

\begin{figure*}[!ht]
\centering
\includegraphics[width=1.8\columnwidth]{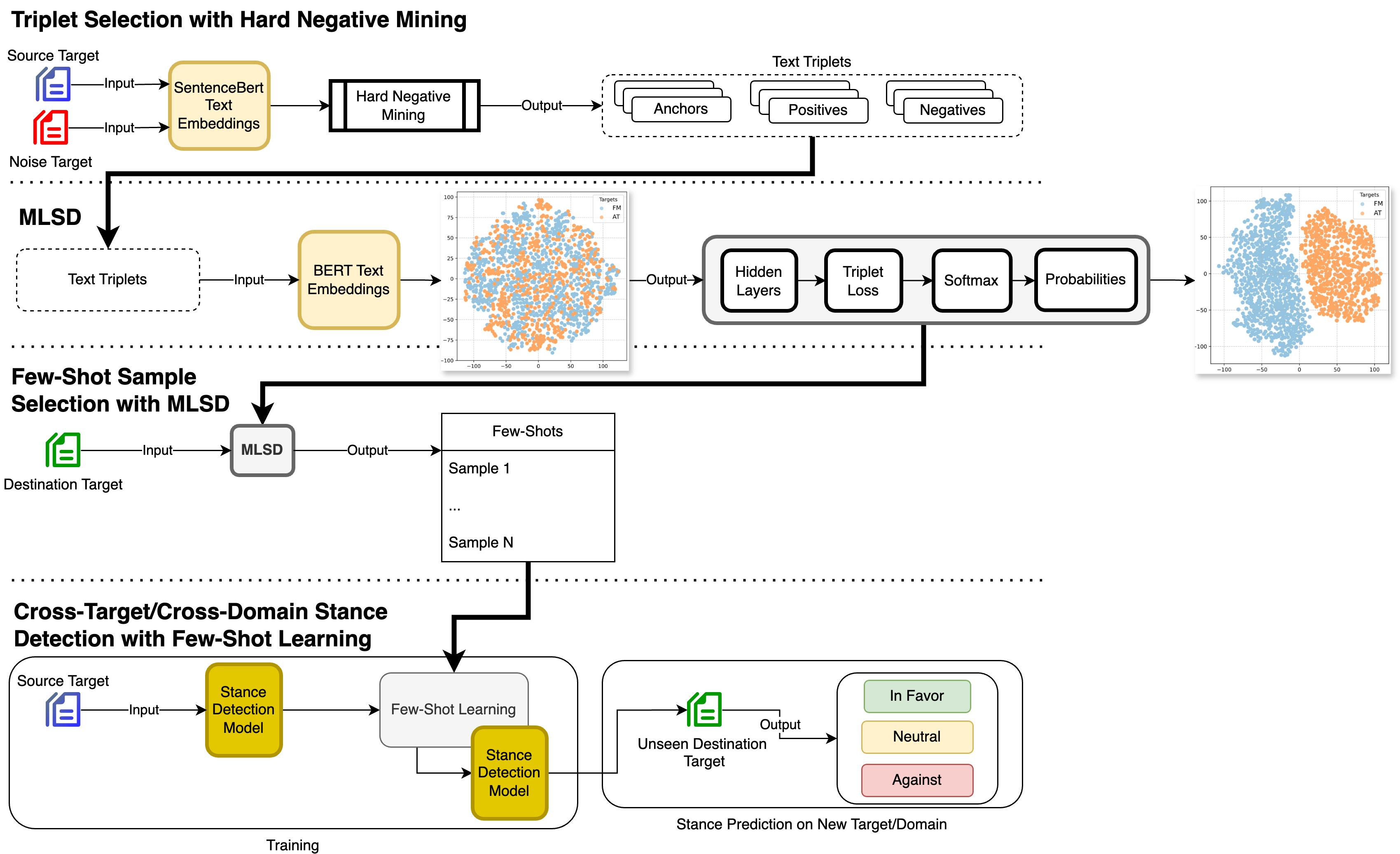}
\caption{Step-by-step workflow of the proposed MLSD architecture. During MLSD, a model is trained to distinguish between data from the source target and another arbitrary noise target, as depicted by the t-SNE visualization of the BERT embeddings for the targets Feminist Movement (FM) and Atheism (AT).}
\label{mlsd}
\end{figure*}


MLSD comprises several steps, including training a similarity detection model using triplet-based metric learning and hard negative mining, identifying the top-$n$ few-shot samples using the trained similarity detection model, and applying few-shot learning using the selected samples for CTSD or CDSD. The  architecture is given in Figure \ref{mlsd}.

\subsection{Step 1: Triplet Selection}

The MLSD approach leverages the similarity between the text embeddings of the source target and destination target data. The idea is to extract a small set of $n$ semantically-similar samples from the destination dataset of size $Z$, where $n \ll Z$, for few-shot learning in CTSD or CDSD. Further, a stance classification model trained on the source target is fine-tuned on these $n$ selected samples and then used to predict the stance on unseen destination data. To enhance the learning of this similarity model, we apply triplet-based metric learning, incorporating data from a noise target.

We first identify the source ($s$) and destination ($d$) targets of interest (e.g., $s=$ Hillary Clinton, $d=$ Donald Trump). We also identify a noise ($n_{\text{noise}}$) target which we define as having no contextually similarity with either $s$ or $d$ (e.g., $n_{\text{noise}}$ = Atheism). We incorporate metric learning, which is a type of machine learning that focuses on learning a similarity metric between data points. The goal of the triplet-based metric learning is to learn an embedding space where examples from similar targets are closer together, while examples from different targets are pushed apart \cite{metric_csur, Ghojogh2023}.  To achieve this, we construct triplets consisting of an anchor from the source target, positive from source target (different example from source target), and negative from noise target.

 Hard negative mining is incorporated to further improve the effectiveness of metric learning and triplet formation. Instead of selecting random negative samples, we use hard negatives, which are examples from the noise target that are particularly challenging, i.e., they are close to the anchor in the learned embedding space but belong to a different target. To achieve this, we leverage SBERT (Sentence-BERT \cite{reimers-2019-sentence-bert}), which provides semantically rich embeddings optimized for similarity tasks, mapping similar text instances closer together in the embedding space.

The process works as follows: for each anchor, we compute its cosine similarity with all candidate negatives in the embedding space. We then sort candidate negatives by similarity in decending order and select one randomly from the top-$k$ most similar negatives as the hard negative, where $k=5$. This ensures that selected hard negative is closer to the anchor in the embedding space, yet belongs to a different target, making is challenging to distinguish. This forces the model to learn a more discriminative representation, improving the ability to generalize across targets and domains.

We create groups of three examples, called triplets, for each text sample. Each triplet includes one example from the source (called the anchor), another example from source (the positive), and a different example from the noise (the negative). The anchor remains the same across all five triplets for each text sample, the positives are chosen randomly, and the negatives are chosen using hard negative mining. We chose to create five triplets per anchor after experimenting with different numbers (5, 10, and 15) and finding it to be the most effective. The goal of creating these triplets is to help the model learn to identify which examples are similar and which are different, so it can better distinguish between different types of text across targets and domains. By keeping the number of triplets low, we avoid repeating too many of the same pairs, which helps prevent the model from becoming too focused on specific patterns and ensures it can generalize well to new data.

Table \ref{metric_table} shows the performance of the metric learning model in terms of accuracy of binary classification between source and negative target, across various target and domain combinations, on their respective test sets. The results shows that the metric learning approach enables models to achieve consistently high accuracy across all target pairs. 

\begin{table}
\centering
\resizebox{.65\columnwidth}{!}{
\begin{tabular}{c|c|c}
\hline
\textbf{Source Target} & \textbf{Noise Target} & \textbf{Accuracy} \\ \hline
FM                     & AT                    & 93.86\%             \\ \hline
LA                     & AT                    & 86.80\%             \\ \hline
HC                     & AT                    & 94.56\%             \\ \hline
DT                     & AT                    & 91.43\%             \\ \hline
ENT                    & POL                   & 99.80\%              \\ \hline
HLT                    & POL                   & 99.15\%             \\ \hline
\end{tabular}}
\caption{Accuracy of the metric learning model in classifying source and noise targets. The model uses embeddings generated from BERT and trained through triplet loss to learn discriminative features.}
\label{metric_table}
\end{table}

\subsection{Triplet Embedding Generation Using BERT}

Once the text triplets are identified, we utilize BERT \cite{devlin-etal-2019-bert} to generate text embeddings for the text triplets. Given a text sequence, $x$, the BERT model is used to produce an embedding vector i.e, $ e(x) = \text{BERT}(x)$, where $e(x) \in \mathbb{R}^d$ represents the embedding of the input sequence, and $d$ is the dimensionality of the embedding space.

Each triplet is denoted as $(A, P, N)$. The triplet loss function $\mathcal{L}_{\text{triplet}}$ is used to ensure that the distance between the anchor and the positive is smaller than the distance between the anchor and the negative by a margin.
\[
\begin{aligned}
\mathcal{L}_{\text{triplet}} = \max(&0, \, d(e(A), e(P)) \\
&- d(e(A), e(N)) + m)
\end{aligned}
\]

where:

\begin{itemize}[noitemsep, topsep=0pt]
    \item $d$ is the Euclidean distance.
    \item $m$ is the margin, a hyperparameter that ensures a minimum separation between similar and dissimilar samples.
\end{itemize}

For discriminating between the source target and another target using the embedded triplets, we develop a deep learning model (i.e., MLSD) using a learning rate of 5e-5 and the Adam optimizer. The margin for the triplet loss was set to 1.0, determined through a grid search to optimize performance. The batch size was 64. The model was trained for 10 epochs, with early stopping based on validation loss to prevent overfitting.


\subsection{Step 2: Top-$N$ Few-Shot Sample Selection}

Once the embedding space is learned and the MLSD model is trained, we use the model to determine which samples from the destination target are most suitable for fine-tuning. Essentially, the MLSD similarity model is used to identify samples from the destination target that are predicted as belonging to the source target, aiding target/domain adaptation. The selection process is based on the classification output from the model, combined with confidence scores and diversity considerations, as follows. 

\begin{enumerate}[noitemsep, topsep=0pt]
    \item Use the trained model to make predictions on the destination dataset $D = \{d_1, d_2, \ldots, d_n\}$.
    \item Compute the confidence scores for each instance using the softmax output from the classifier layer. Let $f_{\theta}(d_i)$ denote the probability for instance $d_i$.
    \item For each stance class $c$, select the top $n$ instances for which the model is most confident:
    \begin{equation}
        \text{Top-N}(c) = \arg \max_{i = 1}^{N} \sum_{d_i \in D_c} f_{\theta}(d_i)
    \end{equation}
    \end{enumerate}

\subsection{Step 3: Classification}

Finally, we train a stance classifier model on a source target. To adapt this model to a different destination target, we fine-tune it using the few-shot samples selected in the previous step. The few-shot samples are assumed to provide high-quality, representative information about the destination target, thereby facilitating effective domain adaptation.

\section{Experimentation}
\label{sec:experiments}

For evaluation, we conducted CTSD and CDSD experiments using two datasets and six stance detection models. We also compared CTSD and CDSD performance with MLSD against the vanilla approach (i.e., training on a source target and testing on a destination target) and CTSD/CDSD with random selection of few-shot samples.

\subsection{Datasets} \label{Datasets}

We used two datasets, including the SemEval 2016 Task 6 Stance Detection  \cite{mohammad2016semeval} and Will-They-Won't-They (WT-WT) \cite{conforti2020willtheywontthey} datasets (see Tables \ref{data-semeval} and \ref{data-wtwt}). The SemEval-2016 dataset contains 4,870 tweets manually annotated for stance (i.e., Favor, Against, or Neither) and target (i.e., Atheism (AT); Climate Change is Real Concern (CC); Feminist Movement (FM); Hillary Clinton (HC); Legalization of Abortion (LA); and Donald Trump (DT)). 

WT-WT was recently released and is currently the largest dataset for stance detection in the English language \cite{conforti2020willtheywontthey}. The dataset includes tweets discussing mergers and acquisitions between different companies in the healthcare and entertainment industries. Each tweet is labeled with the author's stance (i.e., support, refute, comment, or unrelated) towards the domains of Healthcare (HLT) and Entertainment (ENT). 


\begin{table*}[h]
\centering
\resizebox{.8\textwidth}{!}{
\begin{tabular}{r|rrrr|rrrr@{}}
\bottomrule
                                                \textbf{Target}                         & \multicolumn{4}{l|}{\textbf{Training Data}}                                        & \multicolumn{4}{l}{\textbf{Test Data}}                                         \\ \bottomrule
                                 & \textit{\%Favor}          & \textit{\%Against}        & \textit{\%Neutral}           & \textit{\#Total Train}   & \textit{\%Favor}          & \textit{\%Against}        & \textit{\%Neutral}           & \textit{\#Total Test}    \\ \midrule
                                                       Atheism (AT)                        & 17.93          & 59.26          & 22.81          & 513           & 14.55          & 72.73          & 12.73          & 220           \\
                                                       Climate Change is Real Concern (CC) & 53.67          & 3.80           & 42.53          & 395           & 72.78          & 6.51           & 20.71          & 169           \\
                                                       Feminist Movement (FM)              & 31.63          & 49.40          & 18.98          & 664           & 20.35          & 64.21          & 15.44          & 285           \\
                                                       Hillary Clinton (HC)                & 17.13          & 57.04          & 25.83          & 689           & 15.25          & 58.31          & 26.44          & 295           \\
                                                       Legalization of Abortion (LA)       & 18.53          & 54.36          & 27.11          & 653           & 16.43          & 67.50          & 16.07          & 280       
                                                      \\ 
                                                       Donald Trump (DT)       & 20.94          & 42.26          & 36.80          & 530           & 20.90          & 42.37          & 36.73          & 177        \\ \bottomrule
                                                       \textbf{Total}                 & \textbf{25.10} & \textbf{47.03} & \textbf{27.87} & \textbf{3,444} & \textbf{23.94} & \textbf{55.36} & \textbf{20.70} & \textbf{1,427} \\ \bottomrule
\end{tabular}}
\caption{Data distribution in the SemEval-2016 dataset \cite{mohammad2016semeval}.}
\label{data-semeval}
\end{table*}

\begin{table*}[h]
\centering
\resizebox{.8\textwidth}{!}{
\begin{tabular}{r|rrrrr|rrrrr}
\bottomrule
\textbf{Target} & \multicolumn{4}{l}{\textbf{Training Data}}               &       & \multicolumn{5}{l}{\textbf{Test Data}}                        \\
\bottomrule
                & \textit{\%Support} & \textit{\%Refute} & \textit{\%Comment} & \textit{\%Unrelated} & \textit{Total} & \textit{\%Support} & \textit{\%Refute} & \textit{\%Comment} & \textit{\%Unrelated} & \textit{Total} \\ \midrule
Healthcare      & 16.1      & 11.71     & 37.76     & 34.33        & 22,101 & 1187      & 16.11     & 37.76     & 34.33        & 7,367  \\
Entertainment   & 47.74     & 2.45      & 8.23      & 41.75        & 11,141 & 47.74     & 2.45      & 8.23      & 41.75        & 3,714 \\

\midrule
\textbf{Total}       & \textbf{13.48}      & \textbf{8.55}      & \textbf{41.10}        & \textbf{36.85}  & \textbf{33,242} & \textbf{13.48}       & \textbf{8.55} & \textbf{41.10} & \textbf{36.84} & \textbf{11,081} \\
\bottomrule
\end{tabular}}
\caption{Data distribution in the Will They Won't They dataset \cite{conforti2020willtheywontthey}.}
\label{data-wtwt}
\end{table*}

\subsection{Stance Detection Models}

We demonstrate the effectiveness of MLSD by evaluating it against six widely used models for stance detection \cite{xu-etal-2018-cross, liu2019roberta, kim-2014-convolutional, TAN}. These include four RNN-based models, one CNN-based model, and one BERT-based model for comprehensive comparison across different model architectures.

\textbf{BiLSTM}: A BiLSTM (Bidirectional Long Short-Term Memory) encodes text sequences from both forward and backward directions, capturing contextual information from both sides. The hidden states from each direction are then combined to represent the entire sequence \cite{650093}.

\textbf{BiCond}: Utilizes a BiLSTM to encode targets and uses its final states as the initial states of another BiLSTM to encode the text sequence \cite{augenstein2016stance}.

\textbf{CrossNet}: A BiLSTM-based model designed for CTSD, which incorporates an aspect attention layer to focus on relevant parts of the text that are specific to a given target \cite{xu-etal-2018-cross}.

\textbf{TAN}: A model combining a bidirectional RNN and a BiLSTM with a target-specific attention mechanism \cite{TAN}. 

\textbf{TextCNN}: A convolutional neural network designed for sentence classification, utilizing pre-trained word vectors \cite{kim-2014-convolutional}. 

\textbf{RoBERTa}: A transformer-based model derived from BERT, pre-trained on a large corpus using the masked language modeling objective with modifications, such as longer training and removal of the next sentence prediction task \cite{liu2019roberta}.

For each model, we follow the same configuration as outlined in their respective original papers.

\subsection{Training and Testing}

Experimental scenarios consisted of the following:

\begin{itemize}[noitemsep, topsep=0pt]
    \item \textbf{Standard Training}: The model is trained solely on the source target and directly tested on the destination target. No modifications are made.
    
\item \textbf{Few-Shot Learning with Random Selection}: The model is first trained on the source dataset and then fine-tuned using a randomly selected set of $n$-shot samples from the destination dataset's training set. The final evaluation is performed on the destination dataset's test set.
    
\item \textbf{Few-Shot Learning Using MSLD}: The model is trained on the source dataset and then fine-tuned using the top-$n$ few-shot samples from the destination dataset, where $n =$ 5, 10, or 15. These top-$n$ examples are selected using the proposed metric learning-based approach, identified by their highest confidence scores (logits), indicating alignment with the stance classes based on the learned embeddings. For comparison, in the Random Selection setting, the same number of $n$ samples are randomly selected for each stance class.\end{itemize}


\subsection{Source, Destination, and Noise Targets}
We consider various combinations of source and noise targets, including Feminist Movement (FM) and Atheism (AT), Legalization of Abortion (LA) and Atheism (AT), Hilary Clinton (HC) and Atheism (AT), and Donald Trump (DT) and Atheism (AT). To assess CDSD, we assess Entertainment (ENT) and Politics (POL) (where POL is a concatenation of HC and DT). For targets in the SemEval dataset, we use the entire training set, but for WT-WT, we include only a balanced subset of 1,200 examples to match the size of the POL set, which also contains approximately 1,200 examples.




\subsection{Evaluation}
We employ the macro-average of the $F_1$-score to equally weight all stance classes. Similar to the SemEval task, we treat the \textit{Neither} class as a class of no interest during evaluation, whereas for WT-WT, we consider all stance classes. All results are reported as mean of $F_1$-score over five fixed seeds, minimizing the effect of random variations from model initialization. 

\section{Results}

\subsection{Cross-Target and Cross-Domain Stance Detection}
Table \ref{fm_la} presents the performance of stance detection models using the macro-average $F_1$-score across the FM and LA targets. In all models, the few-shot samples selected using the MLSD approach consistently outperform those selected randomly. This trend is similarly observed for the HC and DT targets, as shown in Table \ref{hc_dt}. We attribute performance gains to the learned embeddings for selecting the most informative samples for fine-tuning. By focusing on high confidence examples, the model is better able to adapt to the destination target.


In cross-domain settings, for the ENT and HLT targets, the original authors \cite{conforti2020willtheywontthey} of the dataset highlighted poor generalization performance, with reported $F_1$-scores of 37.77 for HLT $\rightarrow$ ENT and 33.62 for ENT $\rightarrow$ HLT. Our results in Table \ref{ent_pol} show a significant improvement in performance when using few-shot samples selected via MLSD. We also experimented with differnent domains in SemEval dataset, where the destination target is contextually different from the source, we trained the models on FM and tested them on HC and DT. The results, presented in Table \ref{fm_hc_dt}, demonstrate that the few-shot samples acquired using MLSD helped all models achieve better performance compared to randomly selected few-shot samples.


\begin{table*}[h]
\centering
\resizebox{0.8\textwidth}{!}{
\begin{tabular}{c|ccc|ccc}
\hline
         & \multicolumn{3}{c|}{\textbf{LA} $\rightarrow$ \textbf{FM}}                                      & \multicolumn{3}{c}{\textbf{FM} $\rightarrow$ \textbf{LA}}                                      \\ \hline
\textbf{Model} &
  {\textbf{Standard Training}} &
  {\textbf{Random Selection}} &
  \textbf{MLSD} &
  {\textbf{Standard Training}} &
  {\textbf{Random Selection}} &
  \textbf{MLSD} \\ \hline
BiCond   & {36.63\%} & 31.84\% & {\textbf{43.3\%}}  & {35.13\%} & 30.94\% & {\textbf{42.67\%}} \\ \hline
BiLSTM   & {35.99\%} & 33.03\% & {\textbf{42.2\%}}  & {38.73\%} & 29.69\% & {\textbf{39.98\%}} \\ \hline
CrossNet & {35.18\%} & 37.01\% & {\textbf{41.61\%}} & {38.69\%} & 37.8\%  & {\textbf{39.72\%}} \\ \hline
Roberta  & {28.79\%} & 30.56\% & {\textbf{32.15\%}} & {27.88\%} & 28.47\% & {\textbf{30.95\%}} \\ \hline
TAN      & {35.68\%} & 35.77\% & {\textbf{39.96\%}} & {36\%}    & 35.54\% & {\textbf{38.89\%}} \\ \hline
TextCNN  & {26.69\%} & 30.07\% & {\textbf{42.38\%}} & {26.16\%} & 32.5\%  & {\textbf{42.11\%}} \\ \hline
\end{tabular}}
\caption{CTSD performance (macro-average $F_1$-score), averaged across five seeds and 5, 10, and 15-shot settings, with S $\rightarrow$ D indicating source and destination targets. The improvement in performance using MLSD over random selection is statistically significant, with $p < 0.05$ for both LA $\rightarrow$ FM and FM $\rightarrow$ LA, based on a paired $t$-test.}
\label{fm_la}
\end{table*}

\begin{table*}[htbp!]
\centering
\resizebox{0.8\textwidth}{!}{
\begin{tabular}{c|ccc|ccc}
\hline
         & \multicolumn{3}{c|}{\textbf{HC} $\rightarrow$ \textbf{DT}}                                      & \multicolumn{3}{c}{\textbf{DT} $\rightarrow$ \textbf{HC}}                                      \\ \hline
\textbf{Model} &
  {\textbf{Standard Training}} &
  {\textbf{Random Selection}} &
  \textbf{MLSD} &
  {\textbf{Standard Training}} &
  {\textbf{Random Selection}} &
  \textbf{MLSD} \\ \hline
BiCond   & {36.35\%} & 31.78\% & {\textbf{42.59\%}} & {36.7\%}  & 30.44\% & {\textbf{43.04\%}} \\ \hline
BiLSTM   & {38.53\%} & 30.77\% & {\textbf{40.64\%}} & {38.15\%} & 31.17\% & {\textbf{41.46\%}} \\ \hline
CrossNet & {35.09\%} & 37.39\% & {\textbf{40.47\%}} & {34.98\%} & 36.17\% & {\textbf{40.84\%}} \\ \hline
Roberta  & {29.21\%} & 30.07\% & {\textbf{31.58\%}} & {29.53\%} & 30.28\% & {\textbf{31.71\%}} \\ \hline
TAN      & {36.41\%} & 35.14\% & {\textbf{37.79\%}} & {35.01\%} & 35.19\% & {\textbf{38.91\%}} \\ \hline
TextCNN  & {26.79\%} & 30.77\% & {\textbf{41.82\%}} & {26.25\%} & 30.41\% & {\textbf{41.73\%}} \\ \hline
\end{tabular}}
\caption{CTSD performance (macro-average $F_1$-score), averaged across five seeds and 5, 10, and 15-shot settings, with S $\rightarrow$ D indicating source and destination targets. The improvement in performance using MLSD over random selection is statistically significant, with $p < 0.05$ for both HC $\rightarrow$ DT and DT $\rightarrow$ HC, based on a paired $t$-test.}
\label{hc_dt}
\end{table*}

\begin{table*}[htbp!]
\centering
\resizebox{.8\textwidth}{!}{
\begin{tabular}{c|ccc|ccc}
\hline
         & \multicolumn{3}{c|}{\textbf{FM} $\rightarrow$ \textbf{HC}}                                      & \multicolumn{3}{c}{\textbf{FM} $\rightarrow$ \textbf{DT}}                                      \\ \hline
\textbf{Model} &
  {\textbf{Standard Training}} &
  {\textbf{Random Selection}} &
  \textbf{MLSD} &
  {\textbf{Standard Training}} &
  {\textbf{Random Selection}} &
  \textbf{MLSD} \\ \hline
BiCond   & {28.51\%} & 35.67\% & {\textbf{44.84\%}} & {34}    & 28.35 & {\textbf{40.71\%}} \\ \hline
BiLSTM   & {33.45\%} & 27.01\% & {\textbf{47.86\%}} & {34.86\%} & 29.19\% & {\textbf{35.02\%}} \\ \hline
CrossNet & {30.05\%} & 30.57\% & {\textbf{45.6\%}}  & {32.08\%} & 29.66\% & {\textbf{38.27\%}} \\ \hline
Roberta  & {29.2\%}  & 32.85\% & {\textbf{34.62\%}} & {31.76\%} & 31.25\% & {\textbf{34.28\%}} \\ \hline
TAN      & {33.37\%} & 32.93\% & {\textbf{42.76\%}} & {36.34\%} & 32.23\% & {\textbf{36.75\%}} \\ \hline
TextCNN  & {22.25\%} & 28.63\% & {\textbf{42.35\%}} & {23.79\%} & 28.35\% & {\textbf{36.79\%}} \\ \hline
\end{tabular}}
\caption{CDSD performance (macro-average $F_1$-score), averaged across five seeds and 5, 10, and 15-shot settings, with S $\rightarrow$ D indicating source and destination targets. The improvement in performance using MLSD over random selection is statistically significant, with $p < 0.05$ for both FM $\rightarrow$ HC and FM $\rightarrow$ DT, based on a paired $t$-test.}
\label{fm_hc_dt}
\end{table*}

\begin{table*}[htbp!]
\centering
\resizebox{.8\textwidth}{!}{
\begin{tabular}{c|ccc|ccc}
\hline
         & \multicolumn{3}{c|}{\textbf{ENT} $\rightarrow$ \textbf{HLT}}                                    & \multicolumn{3}{c}{\textbf{HLT} $\rightarrow$ \textbf{ENT}}                                    \\ \hline
\textbf{Model} &
  {\textbf{Standard Training}} &
  {\textbf{Random Selection}} &
  \textbf{MLSD} &
  {\textbf{Standard Training}} &
  {\textbf{Random Selection}} &
  \textbf{MLSD} \\ \hline
BiCond   & {29.84\%} & 22.36\% & {\textbf{71.92\%}} & {36.46\%} & 28.55\% & {\textbf{68.92\%}} \\ \hline
BiLSTM   & {31.43\%} & 34.19\% & {\textbf{71.51\%}} & {35.8\%}  & 30.95\% & {\textbf{68.49\%}} \\ \hline
CrossNet & {36.13\%} & 39.3\%  & {\textbf{70.79\%}} & {33.71\%} & 33.29\% & {\textbf{67.89\%}} \\ \hline
Roberta  & {21.51\%} & 24.26\% & {\textbf{24.94\%}} & {20.67\%} & 23.23\% & {\textbf{24.06\%}} \\ \hline
TAN      & {28.66\%} & 32.87\% & {\textbf{62.82\%}} & {34.63\%} & 31.1\%  & {\textbf{62.91\%}} \\ \hline
TextCNN  & {29.48\%} & 27.81\% & {\textbf{66.14\%}} & {30.82\%} & 25.64\% & {\textbf{60.31\%}} \\ \hline
\end{tabular}}
\caption{CDSD performance (macro-average $F_1$-score),  averaged across five seeds and 5, 10, and 15-shot settings, with S $\rightarrow$ D indicating source and destination targets. The improvement in performance using MLSD over random selection is statistically significant, with $p < 0.05$ for both ENT $\rightarrow$ HLT and HLT $\rightarrow$ ENT, based on a paired $t$-test.}
\label{ent_pol}
\end{table*}

\subsection{Results Across Classifiers}

Across all classifiers, MLSD demonstrated notable improvements over the random few shot selection. On average, the performance increased by 11.72\% for FM $\rightarrow$ HC, 7.20\% for FM $\rightarrow$ DT, 31.32\% for ENT $\rightarrow$ HLT, 29.87\% for HLT $\rightarrow$ ENT, 7.27\% for LA $\rightarrow$ FM, 6.56\% for FM $\rightarrow$ LA, 6.47\% for HC $\rightarrow$ DT, and 7.33\% for DT $\rightarrow$ HC, with these values averaged across all classifiers.

An interesting observation is that the RoBERTa model did not achieve as significant an improvement compared to the CNN and RNN based models. This may be due to RoBERTa’s reliance on large-scale pre-training, which already captures much of the generalization required for stance detection, leading to smaller gains from the MLSD approach. In contrast, CNN and RNN models, which do not benefit from such extensive pre-training, gain more from the enhanced sample selection process of MLSD, improving their ability to generalize across both cross-target and cross-domain settings.

\subsection{Few-Shot Learning}

We evaluated 5, 10, and 15 samples for few-shot learning with MLSD. The average performance across all values is shown in Table \ref{across_top-N}. It is noteworthy that even when selecting the maximum number of few-shots, i.e., 15 samples, the proportion of data utilized remains remarkably small. Specifically, we use only 0.02\% of the data from FM, LA, HC each and 0.03\% from DT, 0.001\% from ENT, and an incredibly small 0.0006\% from HLT. These results suggest then a Top N > 5 often provides better results.  

\begin{table}[H]
\resizebox{\columnwidth}{!}{
\begin{tabular}{c|ccc|ccc}
\hline
 & \multicolumn{3}{c|}{\textbf{Cross-Target}}                                       & \multicolumn{3}{c}{\textbf{Cross-Domain}}                                       \\ \hline
\textbf{Model}   & \multicolumn{1}{c|}{\textbf{5}}     & \multicolumn{1}{c|}{\textbf{10}}    & \textbf{15}    & \multicolumn{1}{c|}{\textbf{5}}     & \multicolumn{1}{c|}{\textbf{10}}    & \textbf{15}    \\ \hline
       
BiCond  & \multicolumn{1}{c|}{42.45\%} & \multicolumn{1}{c|}{\textbf{43.37\%}} & 42.85\% & \multicolumn{1}{c|}{56.14\%} & \multicolumn{1}{c|}{56.36\%} & \textbf{57.28\%} \\ \hline
BiLSTM  & \multicolumn{1}{c|}{39.56\%} & \multicolumn{1}{c|}{\textbf{41.84\%}} & 41.8\%  & \multicolumn{1}{c|}{55.81\%} & \multicolumn{1}{c|}{\textbf{55.97\%}} & 55.36\% \\ \hline
CrossNet & \multicolumn{1}{c|}{39.19\%} & \multicolumn{1}{c|}{41.39\%} & \textbf{41.47\%} & \multicolumn{1}{c|}{\textbf{56.09\%}} & \multicolumn{1}{c|}{55.25\%} & 55.57\% \\ \hline
Roberta  & \multicolumn{1}{c|}{30.97\%} & \multicolumn{1}{c|}{31.83\%} & \textbf{31.97\%} & \multicolumn{1}{c|}{29.47\%} & \multicolumn{1}{c|}{29.47\%} & \textbf{29.48\%} \\ \hline
TAN     & \multicolumn{1}{c|}{37.47\%} & \multicolumn{1}{c|}{39.31\%} & \textbf{39.86\%} & \multicolumn{1}{c|}{52.12\%} & \multicolumn{1}{c|}{\textbf{51.48\%}} & 50.31\% \\ \hline
TextCNN & \multicolumn{1}{c|}{41.68\%} & \multicolumn{1}{c|}{42.12\%} & \textbf{42.21\%} & \multicolumn{1}{c|}{51.39\%} & \multicolumn{1}{c|}{\textbf{51.4\%}}  & 51.39\% \\ \hline
\end{tabular}}
\caption{Average stance detection performance (macro-average $F_1$-score) across classifiers for 5, 10, and 15 ``few-shot'' samples in MLSD.}
\label{across_top-N}
\end{table}

\subsection{Comparison with Existing Models}

Across all classifiers, MLSD demonstrated notable performance improvements for CTSD and CDSD over existing models. On average, the performance increased by 13.53\% for FM $\rightarrow$ HC, 4.83\% for FM $\rightarrow$ DT, 31.84\% for ENT $\rightarrow$ HLT, 22.92\% for HLT $\rightarrow$ ENT, 7.10\% for LA $\rightarrow$ FM, 5.28\% for FM $\rightarrow$ LA, 5.41\% for HC $\rightarrow$ DT, and 6.17\% for DT $\rightarrow$ HC, with these values averaged across all classifiers. These results shows that MLSD is highly effective in enhancing stance detection performance, particularly in cross-domain scenarios where traditional methods often struggle.

\section{Conclusion}
\label{sec:conclusion}
This paper introduces MLSD, a novel approach to few-shot learning for CTSD and CDSD. By utilizing metric learning with triplet loss and hard negative mining, MLSD identifies the most informative few-shot samples from a destination target for refining a pre-trained model, significantly improving model performance across multiple stance detection models. Our experiments demonstrated that MLSD consistently outperforms random selection in both cross-target and cross-domain settings, achieving superior results even when the source and destination targets are contextually dissimilar.


\section{Limitations}
\label{sec:limitations}

The effectiveness of metric learning and triplet loss relies heavily on the quality of the precomputed embeddings, which may vary depending on the pre-trained model used (e.g., BERT or RoBERTa). While improvements in the underlying embeddings could further enhance performance, they may also introduce variability across different models. 

Second, the process of hard negative mining adds computational complexity, as it requires calculating similarity scores for numerous candidate pairs, particularly in large datasets. This increases training time, especially as the number of candidates grows. While we limit the number of triplets per anchor to mitigate overfitting, the selection of this number is somewhat heuristic and may require further tuning to generalize across different datasets and tasks.

Finally, MLSD requires labeled few-shot samples from the destination target. Although our method efficiently selects the most informative few-shot examples, it still relies on the availability of labeled data, which can be scarce or expensive to obtain in certain real-world scenarios.

\section{Ethical Impact Statement} \label{Ethical}

This research introduces a metric learning-based few-shot approach (MLSD) for improving stance detection across targets and domains. While effective, it raises ethical considerations, particularly the risk of bias amplification. If training data lacks diversity, the model could reinforce existing biases, leading to unfair outcomes. To counteract this, we prioritize balanced and diverse datasets to ensure broad representation. The application of stance detection in sensitive areas, such as politics or healthcare, also presents risks, including opinion monitoring and automated censorship. It is crucial that the technology is deployed transparently and responsibly to avoid misuse. Lastly, by leveraging few-shot learning, MLSD reduces reliance on large datasets, promoting a more sustainable approach to stance detection.

\bibliography{custom}

\begin{thebibliography}{25}
\providecommand{\natexlab}[1]{#1}

\bibitem[{Arakelyan et~al.(2023)Arakelyan, Arora, and Augenstein}]{arakelyan-etal-2023-topic}
Erik Arakelyan, Arnav Arora, and Isabelle Augenstein. 2023.
\newblock \href {https://doi.org/10.18653/v1/2023.acl-long.752} {Topic-guided sampling for data-efficient multi-domain stance detection}.
\newblock In \emph{Proceedings of the 61st Annual Meeting of the Association for Computational Linguistics (Volume 1: Long Papers)}, pages 13448--13464, Toronto, Canada. Association for Computational Linguistics.

\bibitem[{Augenstein et~al.(2016)Augenstein, Rockt{\"a}schel, Vlachos, and Bontcheva}]{augenstein2016stance}
Isabelle Augenstein, Tim Rockt{\"a}schel, Andreas Vlachos, and Kalina Bontcheva. 2016.
\newblock Stance detection with bidirectional conditional encoding.
\newblock \emph{arXiv preprint arXiv:1606.05464}.

\bibitem[{Conforti et~al.(2020)Conforti, Berndt, Pilehvar, Giannitsarou, Toxvaerd, and Collier}]{conforti2020willtheywontthey}
Costanza Conforti, Jakob Berndt, Mohammad~Taher Pilehvar, Chryssi Giannitsarou, Flavio Toxvaerd, and Nigel Collier. 2020.
\newblock \href {https://doi.org/10.18653/v1/2020.acl-main.157} {Will-they-won{'}t-they: A very large dataset for stance detection on {T}witter}.
\newblock In \emph{Proceedings of the 58th Annual Meeting of the Association for Computational Linguistics}, pages 1715--1724, Online. Association for Computational Linguistics.

\bibitem[{Devlin et~al.(2019)Devlin, Chang, Lee, and Toutanova}]{devlin-etal-2019-bert}
Jacob Devlin, Ming-Wei Chang, Kenton Lee, and Kristina Toutanova. 2019.
\newblock \href {https://doi.org/10.18653/v1/N19-1423} {{BERT}: Pre-training of deep bidirectional transformers for language understanding}.
\newblock In \emph{Proceedings of the 2019 Conference of the North {A}merican Chapter of the Association for Computational Linguistics: Human Language Technologies, Volume 1 (Long and Short Papers)}, pages 4171--4186, Minneapolis, Minnesota. Association for Computational Linguistics.

\bibitem[{Du et~al.(2017)Du, Xu, He, and Gui}]{TAN}
Jiachen Du, Ruifeng Xu, Yulan He, and Lin Gui. 2017.
\newblock \href {https://doi.org/10.24963/ijcai.2017/557} {Stance classification with target-specific neural attention networks}.
\newblock In \emph{26th International Joint Conference on Artificial Intelligence, IJCAI 2017}, pages 3988--3994. International Joint Conferences on Artificial Intelligence.
\newblock IJCAI International Joint Conference on Artificial Intelligence2017, Pages 3988-399426th International Joint Conference on Artificial Intelligence, IJCAI 2017; Melbourne; Australia; 19 August 2017 through 25 August 2017; Code 130864 ; 26th International Joint Conference on Artificial Intelligence, IJCAI 2017 ; Conference date: 19-08-2017 Through 25-08-2017.

\bibitem[{Gera and Neal(2022)}]{gera-neal-2022-comparative}
Parush Gera and Tempestt Neal. 2022.
\newblock \href {https://doi.org/10.18653/v1/2022.eval4nlp-1.7} {A comparative analysis of stance detection approaches and datasets}.
\newblock In \emph{Proceedings of the 3rd Workshop on Evaluation and Comparison of NLP Systems}, pages 58--69, Online. Association for Computational Linguistics.

\bibitem[{Ghojogh et~al.(2023)Ghojogh, Crowley, Karray, and Ghodsi}]{Ghojogh2023}
Benyamin Ghojogh, Mark Crowley, Fakhri Karray, and Ali Ghodsi. 2023.
\newblock \href {https://doi.org/10.1007/978-3-031-10602-6_19} {\emph{Deep Metric Learning}}, pages 531--562.
\newblock Springer International Publishing, Cham.

\bibitem[{Hardalov et~al.(2021)Hardalov, Arora, Nakov, and Augenstein}]{hardalov-etal-2021-cross}
Momchil Hardalov, Arnav Arora, Preslav Nakov, and Isabelle Augenstein. 2021.
\newblock \href {https://doi.org/10.18653/v1/2021.emnlp-main.710} {Cross-domain label-adaptive stance detection}.
\newblock In \emph{Proceedings of the 2021 Conference on Empirical Methods in Natural Language Processing}, pages 9011--9028, Online and Punta Cana, Dominican Republic. Association for Computational Linguistics.

\bibitem[{Jiang et~al.(2022)Jiang, Gao, Shen, and Cheng}]{prompt_based_stance}
Yan Jiang, Jinhua Gao, Huawei Shen, and Xueqi Cheng. 2022.
\newblock \href {https://doi.org/10.1145/3477495.3531979} {Few-shot stance detection via target-aware prompt distillation}.
\newblock In \emph{Proceedings of the 45th International ACM SIGIR Conference on Research and Development in Information Retrieval}, SIGIR '22, page 837–847, New York, NY, USA. Association for Computing Machinery.

\bibitem[{Khiabani and Zubiaga(2023)}]{10098547}
Parisa~Jamadi Khiabani and Arkaitz Zubiaga. 2023.
\newblock \href {https://doi.org/10.1109/TCSS.2023.3264114} {Few-shot learning for cross-target stance detection by aggregating multimodal embeddings}.
\newblock \emph{IEEE Transactions on Computational Social Systems}, pages 1--10.

\bibitem[{Kim(2014)}]{kim-2014-convolutional}
Yoon Kim. 2014.
\newblock \href {https://doi.org/10.3115/v1/D14-1181} {Convolutional neural networks for sentence classification}.
\newblock In \emph{Proceedings of the 2014 Conference on Empirical Methods in Natural Language Processing ({EMNLP})}, pages 1746--1751, Doha, Qatar. Association for Computational Linguistics.

\bibitem[{Li et~al.(2023)Li, Zhao, and Caragea}]{10.1145/3543507.3583250}
Yingjie Li, Chenye Zhao, and Cornelia Caragea. 2023.
\newblock \href {https://doi.org/10.1145/3543507.3583250} {Tts: A target-based teacher-student framework for zero-shot stance detection}.
\newblock In \emph{Proceedings of the ACM Web Conference 2023}, WWW '23, page 1500–1509, New York, NY, USA. Association for Computing Machinery.

\bibitem[{Liang et~al.(2021)Liang, Fu, Gui, Yang, Du, He, and Xu}]{10.1145/3442381.3449790}
Bin Liang, Yonghao Fu, Lin Gui, Min Yang, Jiachen Du, Yulan He, and Ruifeng Xu. 2021.
\newblock \href {https://doi.org/10.1145/3442381.3449790} {Target-adaptive graph for cross-target stance detection}.
\newblock In \emph{Proceedings of the Web Conference 2021}, WWW '21, page 3453–3464, New York, NY, USA. Association for Computing Machinery.

\bibitem[{Liu et~al.(2022)Liu, Lin, Ji, Li, Fu, and Wang}]{semi_supervised_stance}
Rui Liu, Zheng Lin, Huishan Ji, Jiangnan Li, Peng Fu, and Weiping Wang. 2022.
\newblock \href {https://aclanthology.org/2022.coling-1.605} {Target really matters: Target-aware contrastive learning and consistency regularization for few-shot stance detection}.
\newblock In \emph{Proceedings of the 29th International Conference on Computational Linguistics}, pages 6944--6954, Gyeongju, Republic of Korea. International Committee on Computational Linguistics.

\bibitem[{Liu et~al.(2021)Liu, Lin, Tan, and Wang}]{commonsense_stance}
Rui Liu, Zheng Lin, Yutong Tan, and Weiping Wang. 2021.
\newblock Enhancing zero-shot and few-shot stance detection with commonsense knowledge graph.
\newblock In \emph{Findings of the Association for Computational Linguistics: ACL-IJCNLP 2021}, pages 3152--3157.

\bibitem[{Liu(2019)}]{liu2019roberta}
Yinhan Liu. 2019.
\newblock Roberta: A robustly optimized bert pretraining approach.
\newblock \emph{arXiv preprint arXiv:1907.11692}.

\bibitem[{Mohammad et~al.(2016)Mohammad, Kiritchenko, Sobhani, Zhu, and Cherry}]{mohammad2016semeval}
Saif Mohammad, Svetlana Kiritchenko, Parinaz Sobhani, Xiaodan Zhu, and Colin Cherry. 2016.
\newblock Semeval-2016 task 6: Detecting stance in tweets.
\newblock In \emph{Proceedings of the 10th International Workshop on Semantic Evaluation (SemEval-2016)}, pages 31--41.

\bibitem[{Reimers and Gurevych(2019)}]{reimers-2019-sentence-bert}
Nils Reimers and Iryna Gurevych. 2019.
\newblock \href {https://arxiv.org/abs/1908.10084} {Sentence-bert: Sentence embeddings using siamese bert-networks}.
\newblock In \emph{Proceedings of the 2019 Conference on Empirical Methods in Natural Language Processing}. Association for Computational Linguistics.

\bibitem[{Santosh et~al.(2019)Santosh, Bansal, and Saha}]{siamnet}
T.~Y.S.S. Santosh, Srijan Bansal, and Avirup Saha. 2019.
\newblock \href {https://doi.org/10.1145/3297001.3297047} {Can siamese networks help in stance detection?}
\newblock In \emph{Proceedings of the ACM India Joint International Conference on Data Science and Management of Data}, CODS-COMAD '19, page 306–309, New York, NY, USA. Association for Computing Machinery.

\bibitem[{Schuster and Paliwal(1997)}]{650093}
M.~Schuster and K.K. Paliwal. 1997.
\newblock \href {https://doi.org/10.1109/78.650093} {Bidirectional recurrent neural networks}.
\newblock \emph{IEEE Transactions on Signal Processing}, 45(11):2673--2681.

\bibitem[{Sultan et~al.(2022)Sultan, Sil, and Florian}]{sultan-etal-2022-overfit}
Md~Arafat Sultan, Avi Sil, and Radu Florian. 2022.
\newblock \href {https://doi.org/10.18653/v1/2022.emnlp-main.247} {Not to overfit or underfit the source domains? an empirical study of domain generalization in question answering}.
\newblock In \emph{Proceedings of the 2022 Conference on Empirical Methods in Natural Language Processing}, pages 3752--3761, Abu Dhabi, United Arab Emirates. Association for Computational Linguistics.

\bibitem[{Wang et~al.(2020)Wang, Yao, Kwok, and Ni}]{metric_csur}
Yaqing Wang, Quanming Yao, James~T. Kwok, and Lionel~M. Ni. 2020.
\newblock \href {https://doi.org/10.1145/3386252} {Generalizing from a few examples: A survey on few-shot learning}.
\newblock \emph{ACM Comput. Surv.}, 53(3).

\bibitem[{Wei and Mao(2019)}]{10.1145/3331184.3331367}
Penghui Wei and Wenji Mao. 2019.
\newblock \href {https://doi.org/10.1145/3331184.3331367} {Modeling transferable topics for cross-target stance detection}.
\newblock In \emph{Proceedings of the 42nd International ACM SIGIR Conference on Research and Development in Information Retrieval}, SIGIR'19, page 1173–1176, New York, NY, USA. Association for Computing Machinery.

\bibitem[{Wen and Hauptmann(2023)}]{conditional_generation_stance}
Haoyang Wen and Alexander Hauptmann. 2023.
\newblock \href {https://doi.org/10.18653/v1/2023.acl-short.127} {Zero-shot and few-shot stance detection on varied topics via conditional generation}.
\newblock In \emph{Proceedings of the 61st Annual Meeting of the Association for Computational Linguistics (Volume 2: Short Papers)}, pages 1491--1499, Toronto, Canada. Association for Computational Linguistics.

\bibitem[{Xu et~al.(2018)Xu, Paris, Nepal, and Sparks}]{xu-etal-2018-cross}
Chang Xu, C{\'e}cile Paris, Surya Nepal, and Ross Sparks. 2018.
\newblock \href {https://doi.org/10.18653/v1/P18-2123} {Cross-target stance classification with self-attention networks}.
\newblock In \emph{Proceedings of the 56th Annual Meeting of the Association for Computational Linguistics (Volume 2: Short Papers)}, pages 778--783, Melbourne, Australia. Association for Computational Linguistics.

\end{thebibliography}

\end{document}